\documentclass[sigconf, balance=False, dvipsnames, 9pt]{acmart}

\usepackage{lipsum}
\usepackage{multirow}
\usepackage{arydshln}
\usepackage{pgfplots}
\usepackage{subcaption}
\usepackage{amssymb}
\usepackage{amsthm}
\usepackage{marvosym}
\usepackage{pifont}
\usepackage{bold-extra}
\usepackage{enumitem}   

\usepackage[skip=5pt]{caption}

\theoremstyle{definition}
\newtheorem{definition}{Definition}[section]
\newtheorem*{definition*}{}

\AtBeginDocument{%
  \providecommand\BibTeX{{%
    \normalfont B\kern-0.5em{\scshape i\kern-0.25em b}\kern-0.8em\TeX}}}

\copyrightyear{2020}
\acmYear{2020}
\setcopyright{acmcopyright}\acmConference[EdgeSys '20]{3rd International Workshop on Edge Systems, Analytics and Networking}{April 27, 2020}{Heraklion, Greece}
\acmBooktitle{3rd International Workshop on Edge Systems, Analytics and Networking (EdgeSys '20), April 27, 2020, Heraklion, Greece}
\acmPrice{15.00}
\acmDOI{10.1145/3378679.3394533}
\acmISBN{978-1-4503-7132-2/20/04}

\begin{document}

\title{LDP-Fed: Federated Learning with Local Differential Privacy}

\author{Stacey Truex, Ling Liu, Ka-Ho Chow, Mehmet Emre Gursoy, Wenqi Wei}
     \affiliation{%
          \institution{Georgia Institute of Technology, Atlanta, GA 30332}
     }


\begin{abstract}
  This paper presents LDP-Fed, a novel federated learning system with a formal privacy guarantee using local differential privacy (LDP). Existing LDP protocols are developed primarily to ensure data privacy in the collection of single numerical or categorical values, such as click count in Web access logs. However, in federated learning model parameter updates are collected iteratively from each participant and consist of high dimensional, continuous values with high precision (10s of digits after the decimal point), making existing LDP protocols inapplicable. To address this challenge in LDP-Fed, we design and develop two novel approaches. First, LDP-Fed's LDP Module provides a formal differential privacy guarantee for the repeated collection of model training parameters in the federated training of large-scale neural networks over multiple individual participants’ private datasets. Second, LDP-Fed implements a suite of selection and filtering techniques for perturbing and sharing select parameter updates with the parameter server. We validate our system deployed with a condensed LDP protocol in training deep neural networks on public data. We compare this version of LDP-Fed, coined CLDP-Fed, with other state-of-the-art approaches with respect to model accuracy, privacy preservation, and system capabilities. 
\end{abstract}



\maketitle

\section{Introduction}\label{sec:intro}
Traditionally, machine learning (ML) algorithms have required that all relevant training data be held by a trusted central party. However, in the age of IoT, data is often generated and captured from distributed edge locations with different ownerships from multiple independent parties. Distributed systems were therefore developed for the distributed training of ML models through cluster nodes with shared data access or capabilities for data sharing with one or a few trusted central master node(s). However, when the edge nodes are owned by independent parties, there may not exist such a centralized point of trust. Furthermore, legal restrictions such as HIPAA~\cite{act1996health}, CCPA~\cite{mathews2018california}, or GDPR~\cite{regulation2016regulation} and business competitiveness may further limit the sharing of sensitive data.

In response, federated learning (FL) has emerged as an attractive collaborative learning infrastructure. In a FL system, data owners (participants) do not need to share raw data with one another or rely on a single trusted entity for distributed training of ML models. Instead, participants collaborate to jointly train a ML model by executing local training algorithms on their own private local data and only sharing model parameters with the parameter server. This parameter server serves as a central aggregator to appropriately aggregate the local parameter updates and then share the aggregated updates with every participant. While FL allows participants to keep their raw data local, recent work has shown it is insufficient in protecting the privacy of the underlying training data from known inference attacks~\cite{nasr2018comprehensive}. Model parameters exchanged during the training process~\cite{nasr2018comprehensive} as well as outputs from the trained model~\cite{shokri2017membership, truex2019demystifying} remain as attack surfaces for privacy leakage. 

Existing solutions to protect FL systems from such privacy attacks require trusted aggregators~\cite{papernot:pate:2018} or heavy cryptographic techniques~\cite{bonawitz2017practical, truex2019hybrid} which do not allow individual participants to define different local privacy guarantees, are insufficient for meaningfully protecting each high dimensional parameter vector against privacy leakage in the presence of high dimensional parameter vectors~\cite{shokri2015privacy, bonawitz2017practical}, or have focused on low dimensional models~\cite{abadi2016deep, wang2019collecting}. 

In this paper, we proposed LDP-Fed, a novel FL system for the joint training of deep neural network (DNN) models under the protection of the formal local differential privacy framework. LDP-Fed allows participants to efficiently train complex models such that each participant is formally protected from privacy inference attacks according to their own locally defined privacy setting. This paper makes two original contributions. First, we develop a federated training approach that can perform LDP-based perturbation on complex model parameter updates according to the local privacy budget while minimizing the overwhelming impact of noise on the joint ML training process. Second, we also present our parameter update sharing method for the selective sharing of model parameter updates at various rounds of the iterative LDP-Fed training process. We evaluate LDP-Fed against state-of-the-art privacy-preserving FL approaches in both accuracy and system features.

\section{Preliminaries}\label{sec:prelims}

\subsection{Deep Neural Network Training}\label{subsec:DL}
Deep neural network (DNN) models are composed of many layers of basic building block nodes such as affine functions or simple non-linear functions (e.g. sigmoids, rectified linear units (ReLU), etc.). A DNN model is therefore trained by fitting the parameters of these nodes to a known set of training inputs (provided to the first layer of nodes) and outputs (desired output from the last layer).

Specifically, a loss function $\mathcal{L}$ is quantifies the error between the desired outputs and the DNN generated output. Given a DNN with parameters $\theta$, the loss $\mathcal{L}(\theta)$ of the DNN on the training set $\{x_1, x_2, \dots, x_N\}$ is the average loss over the set, i.e. $\mathcal{L}(\theta) = \frac{1}{N}\sum_i \mathcal{L}(\theta, x_i)$. DNN training therefore seeks the parameters $\theta$ which minimize this loss. While ideally training will result in the loss global minima, training in practice is rarely expected to reach this global value and instead finds an acceptably small loss point.

The process of minimizing the loss $\mathcal{L}$ is often done through applying the technique known as stochastic gradient descent (SGD) iteratively to subsets of the training data known as minibatches. At each step a batch $B$ is selected and an estimation of the gradient $\nabla_\theta \mathcal{L}(\theta)$ is computed as $\mathbf{g}_B = \frac{1}{|B|}\sum_{x\in B} \nabla_\theta \mathcal{L}(\theta, x)$. The training algorithm then updates $\theta$ in the direction $-\mathbf{g}_B$ toward a local minima. Multiple systems are available to enable efficient training and evaluation of these DNNs models~\cite{ierusalimschy1996lua, collobert2011torch7, abadi2016tensorflow}.

\begin{figure}
    \centering
    \includegraphics[width=\columnwidth]{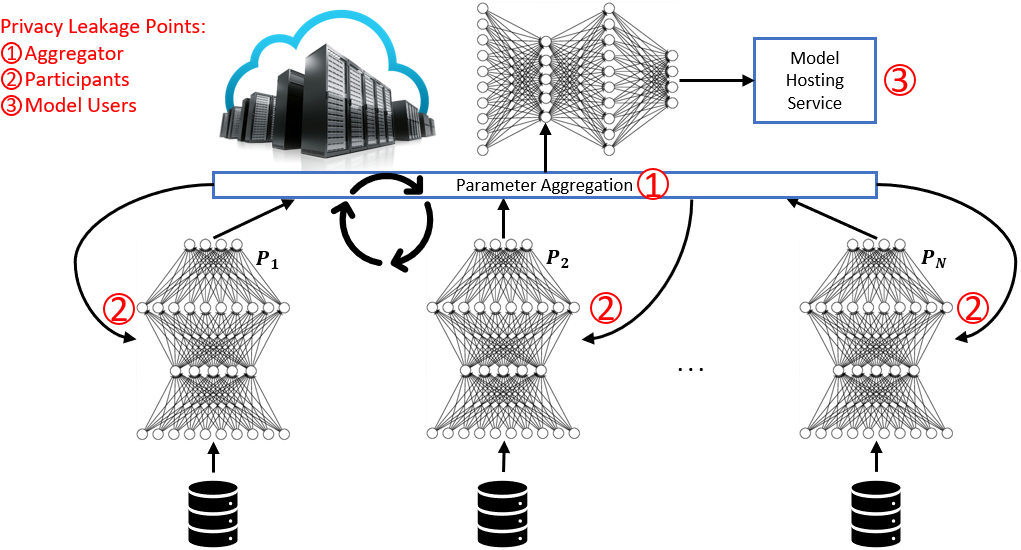}
    \caption{Privacy leakage in federated learning systems.}
    \label{fig:FLleaks}
    \vspace{-\baselineskip}
\end{figure}
\subsection{Federated Learning}\label{subsec:FL}
As privacy concerns and legislation continues to mount, FL systems such as~\cite{bonawitz2019towards} have seen increased attention. FL systems remove the necessity of a central data location to train DNNs. Model parameters which minimize loss across multiple datasets are instead identified through model training that is done locally at the edge.

In a FL setting, $N$ participants, each with independent datasets containing the same features and output classes, agree on a DNN model architecture. A central server (aggregator) then randomly initializes the model parameters $\theta_0$ which are then distributed to each participant so that each may initialize their own copy of the model. At each round $r \in [0, E)$ of training, participants receive a copy of the aggregator's model parameters $\theta_r$. Each participant $P_i$ then conducts model training locally as described in Section~\ref{subsec:DL} to generate updated parameters $\theta_{r+1, i}$ and uploads them to the aggregator. The aggregator then computes the average of value for each parameter and updates the global model with the parameters $\theta_{r+1} = \frac{1}{N}\sum_i \theta_{r+1,i}$. This process is continued either for a pre-determined number of rounds $E$ or until the model converges.

While FL allows for private data to remain local to each participant, this data locality approach proves insufficient in protecting training data privacy as FL systems remain vulnerable to privacy inference attacks. Figure~\ref{fig:FLleaks} highlights the multiple points of potential privacy leakage in federated learning. Information may leak to the central aggregator service (leakage point 1) as well as other participants (leakage point 2) by way of the shared parameter updates which are a type of encoding of each participant's private data. Recent work has indeed demonstrated that effective membership inference privacy attacks may be launched given access to these shared model updates~\cite{nasr2018comprehensive}. Additionally, the final model itself will also leak with prediction outputs (leakage point 3) leading attackers to infer information about the underlying training data points~\cite{shokri2017membership, truex2019demystifying}.
\begin{figure*}[t]
    \centering
    \includegraphics[width=0.75\textwidth]{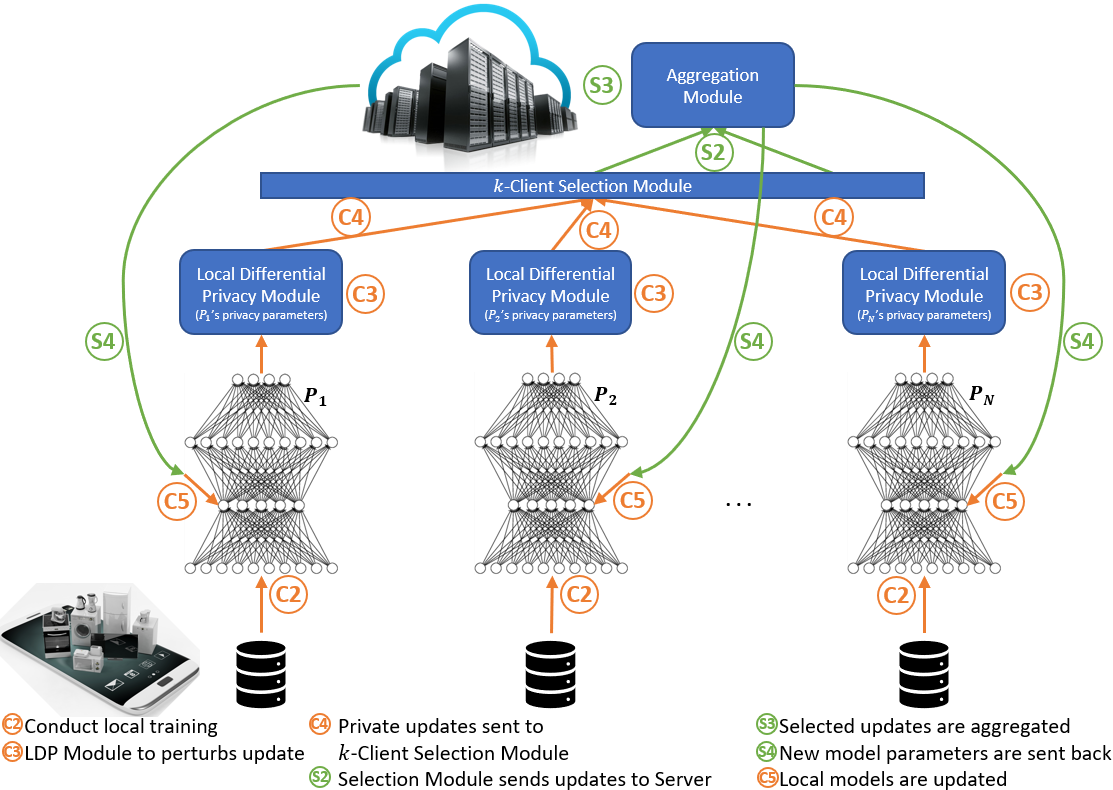}
    \caption{Private Federated Learning with LDP-Fed}
    \label{fig:LDP-Fed}
    \vspace{-\baselineskip}
\end{figure*}

\subsection{Local Differential Privacy}\label{subsec:ldp}
To combat inference attacks against shared data values, companies including Google, Apple, and Microsoft employ local differential privacy (LDP)~\cite{erlingsson2014rappor, thakurta2017emoji, ding2017collecting}, the state-of-the-art in privacy-preserving data collection. Rather than uploading raw data values, users in an LDP system perturb their data $v$ using an algorithm $\Psi$ and instead upload $\Psi(v)$. This perturbed value $\Psi(v)$ is then guaranteed to protect $v$ from inference attacks according to a privacy parameter $\epsilon$ where a lower $\epsilon$ value indicates a higher level of privacy. This guarantee is formalized as follows.
\theoremstyle{definition}
\begin{definition}{($\epsilon$-LDP).}
A randomized algorithm $\Psi$ satisfies $\epsilon$-local differential privacy ($\epsilon$-LDP), where $\epsilon > 0$, if and only if for any inputs $v_1, v_2$ in universe $\mathcal{U}$, we have:
\begin{equation*}
    \forall y \in Range(\Psi): \frac{Pr[\Psi(v_1) = y]}{Pr[\Psi(v_2) = y]} \leq e^\epsilon
\end{equation*}
where Range($\Psi$) is the set of all possible outputs of algorithm $\Psi$.
\end{definition}

\subsubsection{Condensed Local Differential Privacy}
In~\cite{gursoy2019secure}, authors propose a specialization of LDP, Condensed Local Differential Privacy (CLDP). CLDP ensures privacy according to a privacy parameter $\alpha$ where, as with $\epsilon$, a lower $\alpha$ value indicates a higher level of privacy. CLDP, however, also considers a distance metric $d$ in its perturbation algorithm $\Phi$. Specifically, let $\mathcal{U}$ denote the finite universe of possible values for user data $v$. Additionally, let $d : \mathcal{U} \times \mathcal{U} \rightarrow [0, \infty)$ be a distance function that measures the distance between any two items $v_1, v_2 \in \mathcal{U}$. CLDP is then formalized as follows.
\theoremstyle{definition}
\begin{definition}{($\alpha$-CLDP).}
A randomized algorithm $\Phi$ satisfies $\alpha$-condensed local differential privacy ($\alpha$-CLDP), where $\alpha > 0$, if and only if for any inputs $v_1, v_2 \in \mathcal{U}$:
\begin{equation*}
    \forall y \in Range(\Phi) : \frac{Pr[\Phi(v_1) = y]}{Pr[\Phi(v_2) = y]} \leq e^{\alpha\cdot d(v_1, v_2)}
\end{equation*}
where Range($\Phi$) is the set of all possible outputs of algorithm $\Phi$.
\end{definition}

While the definitions of LDP and CLDP are similar, their privacy parameters and indistinguishability properties vary as $\alpha$-CLDP, indistinguishability is also controlled by the items’ distance $d(\cdot, \cdot)$ in addition to $\alpha$. Therefore, as $d$ increases, $\alpha$ must decrease to compensate, making $\alpha \ll \epsilon$. Previous work~\cite{gursoy2019secure} provides details for converting $\epsilon$ to $\alpha$. To guarantee $\alpha$-CLDP, the Exponential Mechanism (EM) is applied to a raw user value $v$ end as follows.

\textbf{Exponential Mechanism (EM)}. Let $v \in \mathcal{U}$ be the raw user data, and let the Exponential Mechanism $\Phi_{EM}$ take as input $v$ and output a perturbed value in $\mathcal{U}$, i.e., $\Phi_{EM} : \mathcal{U} \rightarrow \mathcal{U}$. Then, $\Phi_{EM}$ that produces output $y$ with the following probability satisfies $\alpha$-CLDP:
\begin{equation*}
    \forall y \in \mathcal{U} : Pr[\Phi_{EM}(v) = y] = \frac{e^{\frac{-\alpha\cdot d(v, y)}{2}}}{\sum_{z \in \mathcal{U}}e^{\frac{-\alpha\cdot d(v, z)}{2}}}
\end{equation*}

\subsubsection{Privacy Accounting}
In differentially private federated training of DNN models, an important issue arises in accounting for the multiple iterations of the training algorithm. That is, each participant and therefore each private dataset will be queried during multiple rounds of training. The composability of differential privacy allows for such iterations to be accounted for by accumulating the privacy cost at each round of the training as well as the multiple parameters included in each model update. 

To account for the iterative nature of DNN training, the Sequential Composition theorem states that for functions $f_1, \ldots, f_n$ where $f_i$ satisfies $\epsilon_i$-DP for each $i \in [1, n]$, the release of the outputs $f_1(D), \ldots, f_n(D)$ satisfies $(\sum_{i=1}^n \epsilon_i)$-DP. The privacy amplification theorem~\cite{beimel2014bounds, kasiviswanathan2011can} additionally states that if random samples are selected rather than all available data, then each round satisfying $\epsilon$-DP incurs only a cost of $(q\epsilon)$ against the privacy budget where $q = L/N$ is the sampling ratio.

\section{Federated Learning with LDP-Fed}\label{sec:ldp-fed}
The LDP-Fed system coordinates the federated learning of a DNN with $N$ participants (clients) and one parameter server. LDP-Fed integrates a LDP privacy guarantee into the general architecture of the FL algorithm as shown in Figure~\ref{fig:LDP-Fed} to protect participants' data from inference attacks. 

Specifically, consider $N$ participants with the same dataset structure and learning task who wish to collaboratively train a DNN model in a federated fashion. That is, each participant wishes to perform local training on its own private data and only share parameter updates to the server. Additionally, participants wish to address FL privacy risks (Section~\ref{subsec:FL}) with an individualized LDP guarantee (Section~\ref{subsec:ldp}). To accomplish these goals, we present the federated training process of our system LDP-Fed, from both client (participant) and server perspectives: 

On the individual client side:
\begin{enumerate}
    \item Participants initialize local DNN instances with model parameters $\theta_0$ and each local LDP Module is initialized with privacy parameters according to individual preferences.
    \item \label{step:start_itr} Each participant locally computes training gradients according to their private, local dataset.
    \item Each participant performs perturbation on their gradients according to their local instance of the LDP Module.
    \item Model parameter updates are anonymously sent to the $k$-Client Selection Module which uniformly at random accepts or rejects updates with probability $q=k/N$.
    \item Each participant waits to receive aggregated parameter updates from the parameter server. Upon receiving the aggregated updates, each participant updates its local DNN model, and proceeds to step~\ref{step:start_itr} to start the next iteration.
\end{enumerate}

On the parameter server side:
\begin{enumerate}
    \item The parameter server generates initial model parameters $\theta_0$ and sends to each participant.
    \item The server waits to receive $k$ parameter updates randomly selected by the $k$-Client Selection Module. 
    \item Once parameter updates are received, the Aggregation Module aggregates the updates, i.e. averages the gradient updates to determine new model parameters.
    \item The parameter server updates model parameters and sends updated values back to participants to update local models.
\end{enumerate}
 
The above steps iterate for both the $N$ clients and the parameter sever until a pre-determined condition is reached such as reaching a maximum number of rounds (iterations) or a public test set no longer reporting improved performance (convergence). Compared with traditional FL systems, LDP-Fed introduces two new components: (1) the Local Differential Privacy Module running on each of the $N$ clients and (2) the $k$-Client Selection Module. 

{\bf Local Differential Privacy Module.\/} For each client, the LDP Module takes as input the high dimensional vector of model parameter updates, say 29,034 distinct values, and outputs a vector containing the perturbed updates according to the participant’s chosen privacy context. In the first prototype of LDP-Fed, we set the default privacy definition to be $\alpha$-CLDP-Fed, a variation of $\alpha$-CLDP. While the definition of $\alpha$-CLDP in~\cite{gursoy2019secure} is provided for LDP perturbation on single integer values in finite spaces, gradient values are instead real values with high precision (10s after decimal points). Therefore the $\alpha$-CLDP-Fed Module introduces a precision parameter $\rho$ and a clipping range parameter $c$ such that each parameter update is converted to an integer in the range [$-c\cdot10^\rho$, $c\cdot10^\rho$]. By transforming the clipped parameters into integers according to the precision parameter $\rho$ and clipping range parameter $c$, we can define the $\alpha$-CLDP-Fed system with Ordinal-CLDP using EM from~\cite{gursoy2019secure}. Larger $c$ and $\rho$ values will result in a larger universe space but allow for more specificity in the model update. 

Another problem with applying $\alpha$-CLDP from~\cite{gursoy2019secure} to FL is that its protocol only accounts for single item uploads. In FL, LDP-Fed needs to iteratively upload a high dimensional parameter vector, which has typically 10,000 or more real valued parameters of high precision. Assume $k=N$ in the $k$-Client Selection Module, let $E$ be the total number of iterations for a FL task, and let $\alpha$ be the total privacy budget. To guarantee $\alpha$-CLDP, we must partition $\alpha$ into $E$ small budgets, one for each of the $E$ total iterations such that $\alpha = \sum_{i=0}^{E-1} \alpha_i$. Let $\theta_i$ be the total number of parameter updates to be uploaded to the parameter server at the $i$th iteration from any of the $k$ selected clients, with $\alpha_i$ denoting the portion of the overall privacy budget $\alpha$ allocated to the $i$th iteration. To guarantee privacy in LDP-Fed, we therefore must set $\alpha_p = \frac{\alpha_i}{|\theta_i|}$ as the privacy budget when applying Ordinal-CLDP to each parameter update in $\theta_i$.

{\bf $k$-Client Selection Module.\/} Just as conventional FL systems do not require every participant to share their local training parameter updates in each round, training in LDP-Fed results in only $k$ participants' parameter updates being uploaded to the parameter server for any given round with $k<=N$. As the discarded updates do not introduce any privacy cost, sampling amplification allows for a tighter bound of $\alpha = \sum_{i=0}^{E-1} q\cdot\alpha_i$ with $q =\frac{k}{N} <= 1$.

\section{Experimental Results}\label{sec:exp}
All experiments were conducted on an example FL system with $N=50$ participants and the $k$-client Selection Module set to randomly select $k=9$ updates at each round. The DNN model architecture used has two convolutional layers each followed by a batch normalization layer and a 2D max-pool layer. The final network layer is a single fully connected layer with 1,568 hidden units. We conduct 80 total rounds of training, i.e. $E$ is set to 80. To evaluate the effectiveness of LDP-Fed, we also implemented a number of related methods, such as Non-Private FL, Local Learning, and secure multiparty computation (SMC) methods for comparison and analysis. Related methods requiring $\epsilon$ values were set with the $\epsilon$ value equivalent to $\alpha=1.0$ given the appropriate $\rho$ and $c$ settings according to the conversion approach provided in~\cite{gursoy2019secure}. 

{\bf Non-Private FL\/}. In non-private federated learning, the LDP Module is not activated and the $k$-Client Selection Module receives complete model parameter updates from participants in the clear. 
{\bf Local Learning\/}. The results of local learning are reported as the average accuracy results received by the individual participants if they were to train the DNN model on their own local datasets without sharing parameter updates. 
{\bf Baseline\/}. Random guess baseline of 10\%.
{\bf SMC\/}. With SMC, the same process as Non-Private FL is followed except that model updates are encrypted when sent to the $k$-Client Selection Module and then decrypted only post-aggregation in the Aggregation Module. Here parameter updates again need to be integers and therefore only $\rho=$10 digits after the decimal are preserved. 
{\bf Differentially Private Stochastic Gradient Descent\/} (DPSGD). Authors in~\cite{abadi2016deep} propose a centralized approach to differentially private deep learning wherein noise is added to each gradient by the optimizer. We compare the impact of using LDP-Fed with the impact of using such a differentially private optimizer on each participant. 
{\bf SMC and DPSGD Hybrid\/} (Hybrid-One). Authors in~\cite{truex2019hybrid} propose a FL system which leverages an optimizer similar to that in the DPSGD method. However, the hybrid approach leverages SMC to decrease the scale of noise required at each participant.

All experiments are carried out on the FashionMNIST dataset, consisting of 60,000 training examples and 10,000 testing examples~\cite{xiao2017/online}. Each example is a 28 ×28 size gray-level image depicting an item from one of ten different fashion classes.

\begin{figure}
\resizebox{\columnwidth}{!}{
    \begin{tikzpicture}
    \begin{axis}[
        xlabel={Round},
        ylabel={Accuracy (\%)},
        xmin=0, xmax=80,
        ymin=0, ymax=100,
        xtick={0,10,20,30,40,50, 60, 70, 80},
        ytick={0,20,40,60,80,100},
        legend pos=outer north east,
        ymajorgrids=false,
    ]
            
    \addplot[
        color=black,
        dotted
        ]
        coordinates {
        (1,71.47)(2,66.59)(3,63.16)(4,67.9)(5,63.65)(6,53.32)(7,59.27)(8,86.09)(9,86.56)(10,86.85)
        (11,86.8)(12,87.46)(13,87.91)(14,87.43)(15,87.57)(16,87.68)(17,88.5)(18,87.82)(19,88.58)(20,88.21)
        (21,88.76)(22,88.54)(23,88.58)(24,88.61)(25,88.85)(26,88.58)(27,88.69)(28,88.35)(29,88.96)(30,88.6)
        (31,89.18)(32,89.24)(33,89.07)(34,89.22)(35,89.22)(36,89.28)(37,89.32)(38,89.3)(39,88.94)(40,89.18)
        (41,89.34)(42,89.45)(43,89.59)(44,89.7)(45,89.85)(46,89.74)(47,89.57)(48,89.58)(49,89.45)(50,89.79)
        (51,89.39)(52,89.69)(53,89.72)(54,89.29)(55,89.45)(56,89.69)(57,89.53)(58,89.45)(59,89.94)(60,90.09)
        (61,88.74)(62,89.66)(63,89.15)(64,88.89)(65,89.72)(66,89.59)(67,89.57)(68,89.84)(69,89.69)(70,89.78)
        (71,89.58)(72,89.51)(73,88.52)(74,89.82)(75,89.77)(76,89.73)(77,89.76)(78,89.41)(79,89.99)(80,90.01)
        };
            
    \addplot[
        color=black,
        dashed
        ]
        coordinates {
        (1,71.87)(2,76.48)(3,75.89)(4,79.8)(5,80.51)(6,80.79)(7,83.01)(8,82.52)(9,83.71)(10,83.86)
        (11,83.74)(12,83.75)(13,83.76)(14,83.87)(15,83.78)(16,83.91)(17,83.79)(18,83.86)(19,83.92)(20,83.89)
        (21,83.89)(22,83.89)(23,83.86)(24,83.92)(25,83.87)(26,83.83)(27,83.86)(28,83.86)(29,83.8)(30,83.93)
        (31,83.89)(32,83.88)(33,83.9)(34,83.79)(35,83.88)(36,83.9)(37,83.85)(38,83.86)(39,83.76)(40,83.8)
        (41,83.89)(42,83.86)(43,83.81)(44,83.85)(45,83.9)(46,83.83)(47,83.81)(48,83.94)(49,83.87)(50,83.79)
        (51,83.85)(52,83.85)(53,83.82)(54,83.9)(55,83.83)(56,83.86)(57,83.88)(58,83.86)(59,83.86)(60,83.93)
        (61,83.82)(62,83.86)(63,83.86)(64,83.82)(65,83.89)(66,83.8)(67,83.83)(68,83.87)(69,83.87)(70,83.86)
        (71,83.82)(72,83.8)(73,83.8)(74,83.79)(75,83.8)(76,83.86)(77,83.82)(78,83.79)(79,83.77)(80,83.85)
        };
            
    \addplot[
        color=black,
        dashdotted
        ]
        coordinates {
        (1,74.33)(2,82.22)(3,84.36)(4,85.56)(5,86.19)(6,85.75)(7,85.33)(8,86.18)(9,87.15)(10,87)
        (11,87.56)(12,88.03)(13,87.71)(14,88.01)(15,87.72)(16,88.08)(17,88.06)(18,87.99)(19,88.51)(20,88.35)
        (21,88.59)(22,88.63)(23,88.38)(24,88.67)(25,88.45)(26,88.42)(27,88.86)(28,89.12)(29,89.02)(30,89.11)
        (31,89.17)(32,89.02)(33,89.23)(34,89.48)(35,89.15)(36,89.19)(37,89.15)(38,89.25)(39,89.37)(40,89.36)
        (41,89.47)(42,89.61)(43,89.49)(44,89.71)(45,89.56)(46,89.8)(47,89.52)(48,89.85)(49,89.7)(50,89.92)
        (51,89.78)(52,89.88)(53,89.82)(54,89.82)(55,89.64)(56,89.8)(57,89.99)(58,89.99)(59,89.94)(60,89.85)
        (61,89.84)(62,90.03)(63,89.84)(64,89.91)(65,90.12)(66,89.93)(67,89.86)(68,90.16)(69,90.14)(70,90.05)
        (71,89.93)(72,90.06)(73,89.95)(74,90.01)(75,90.05)(76,90.07)(77,90.17)(78,90.28)(79,90.29)(80,90.24)
        };
            
    \addplot[
        color=olive,
        mark=square,
        mark repeat = 5,
        mark phase = 1,
        ]
        coordinates {
        (1,47.68)(2,53.61)(3,60.98)(4,66.26)(5,70.08)(6,72.08)(7,71.78)(8,70.4)(9,71.82)(10,73.02)
        (11,71.92)(12,71.94)(13,73.7)(14,73.9)(15,75.52)(16,75.86)(17,75.97)(18,76.21)(19,75.97)(20,76.67)
        (21,76.87)(22,77.28)(23,77.36)(24,77.39)(25,76.68)(26,76.5)(27,76.37)(28,76.59)(29,74.88)(30,74.62)
        (31,73.29)(32,74.96)(33,74.07)(34,78.54)(35,78.91)(36,78.33)(37,78.79)(38,78.07)(39,77.75)(40,78.43)
        (41,78.84)(42,77.99)(43,77.06)(44,77.54)(45,77.96)(46,79.58)(47,79.61)(48,79.64)(49,79.21)(50,79.49)
        (51,79.86)(52,79.78)(53,79.87)(54,79.74)(55,79.55)(56,79.89)(57,79.86)(58,79.95)(59,79.75)(60,79.66)
        (61,79.72)(62,80.11)(63,79.96)(64,80.09)(65,79.99)(66,80.19)(67,80.26)(68,80.43)(69,80.36)(70,79.9)
        (71,79.79)(72,79.86)(73,79.77)(74,79.71)(75,79.77)(76,79.78)(77,79.7)(78,79.88)(79,80.06)(80,80.04)
        };
            
    \addplot[
        color=red,
        mark=triangle,
        mark repeat = 5,
        mark phase = 1,
        ]
        coordinates {
        (1,40.13)(2,48.89)(3,53.83)(4,61.43)(5,61.24)(6,65.3)(7,68.76)(8,71.34)(9,71.37)(10,71.64)
        (11,72.71)(12,73.36)(13,73.84)(14,73.07)(15,73.28)(16,73.82)(17,74.42)(18,74.03)(19,74.56)(20,73.97)
        (21,74.14)(22,74.25)(23,75.32)(24,79.59)(25,79.81)(26,79.79)(27,80.12)(28,80.15)(29,80.14)(30,80.61)
        (31,80.78)(32,80.87)(33,80.81)(34,80.82)(35,80.62)(36,80.99)(37,81.13)(38,81.25)(39,81.32)(40,81.5)
        (41,81.34)(42,81.33)(43,81.42)(44,81.86)(45,81.69)(46,81.96)(47,82.03)(48,82.17)(49,82.27)(50,82.35)
        (51,82.47)(52,82.33)(53,82.36)(54,82.45)(55,82.3)(56,82.61)(57,82.41)(58,82.5)(59,82.37)(60,82.31)
        (61,82.44)(62,82.57)(63,82.79)(64,82.7)(65,82.69)(66,82.77)(67,82.7)(68,82.82)(69,82.81)(70,83.09)
        (71,83.04)(72,83.29)(73,83.06)(74,82.97)(75,83.19)(76,83.15)(77,83.39)(78,83.04)(79,83.34)(80,83.39)
        };
            
    \addplot[
        color=blue,
        mark=o,
        mark repeat = 5,
        mark phase = 1,
        ]
        coordinates {
        (1,26.1)(2,22.22)(3,24.02)(4,48.65)(5,52.04)(6,70.06)(7,73.54)(8,73.55)(9,79.33)(10,80.63)
        (11,81.37)(12,81.26)(13,81.2)(14,82.17)(15,82.18)(16,82.1)(17,82.38)(18,82.65)(19,82.88)(20,82.81)
        (21,83.18)(22,83.27)(23,83.41)(24,83.85)(25,83.91)(26,83.93)(27,84.25)(28,84.54)(29,84.55)(30,84.62)
        (31,84.67)(32,84.8)(33,84.83)(34,84.99)(35,84.92)(36,85.11)(37,85.1)(38,85.24)(39,84.84)(40,84.85)
        (41,85)(42,85.1)(43,85.13)(44,85.17)(45,85.26)(46,85.31)(47,85.31)(48,85.36)(49,85.53)(50,85.62)
        (51,85.65)(52,85.88)(53,85.87)(54,85.92)(55,85.93)(56,85.9)(57,86.03)(58,85.98)(59,86.12)(60,86.2)
        (61,86.2)(62,86.54)(63,86.56)(64,86.64)(65,86.5)(66,86.73)(67,86.73)(68,86.79)(69,86.53)(70,86.85)
        (71,86.67)(72,86.67)(73,86.83)(74,86.82)(75,86.86)(76,86.84)(77,86.82)(78,86.85)(79,86.85)(80,86.85)
        };
        
    \addplot[
        color=blue,
        mark=+,
        mark options=solid,
        mark repeat = 5,
        mark phase = 1,
        dashed
        ]
        coordinates {
        (1,30.08)(2,24.66)(3,22.72)(4,23.65)(5,21.23)(6,61.4)(7,73.7)(8,76.94)(9,77.72)(10,77.86)
        (11,77.89)(12,77.9)(13,77.92)(14,78.26)(15,78.42)(16,78.8)(17,79.61)(18,80.44)(19,81.13)(20,81.16)
        (21,81.14)(22,81.15)(23,80.77)(24,80.9)(25,80.74)(26,82.29)(27,82.27)(28,82.03)(29,82.05)(30,82.29)
        (31,82.32)(32,81.88)(33,82.63)(34,83.23)(35,83.42)(36,83.4)(37,83.42)(38,83.44)(39,83.54)(40,83.55)
        (41,83.65)(42,83.75)(43,83.74)(44,83.75)(45,83.76)(46,83.76)(47,83.89)(48,84.26)(49,84.2)(50,84.39)
        (51,84.34)(52,84.35)(53,84.34)(54,84.34)(55,84.44)(56,84.41)(57,84.66)(58,84.77)(59,84.77)(60,84.76)
        (61,84.74)(62,84.76)(63,84.71)(64,84.78)(65,84.84)(66,84.84)(67,84.67)(68,84.67)(69,84.67)(70,84.68)
        (71,84.82)(72,84.85)(73,84.84)(74,84.9)(75,84.91)(76,84.9)(77,84.9)(78,84.93)(79,84.89)(80,84.89)
        };
    
    \addplot[
        color=blue,
        mark=asterisk,
        mark options=solid,
        mark repeat = 5,
        mark phase = 1,
        dotted
        ]
        coordinates {
        (1,7.97)(2,8)(3,6.71)(4,6.86)(5,6.78)(6,6.58)(7,6.92)(8,9.25)(9,7.73)(10,7.15)
        (11,7.54)(12,7.21)(13,7.15)(14,6.82)(15,6.83)(16,5.04)(17,4.19)(18,3.94)(19,9.64)(20,9.64)
        (21,9.67)(22,10.13)(23,12.79)(24,13.51)(25,13.32)(26,13.24)(27,13.66)(28,13.49)(29,13.35)(30,6.12)
        (31,6.38)(32,6.45)(33,6.34)(34,6.64)(35,6.23)(36,5.03)(37,4.82)(38,4.55)(39,4.27)(40,4.74)
        (41,4.06)(42,4.74)(43,5.28)(44,5.8)(45,5.81)(46,5.6)(47,5.95)(48,6.27)(49,6.8)(50,6.03)
        (51,6.27)(52,5.4)(53,4.88)(54,5.01)(55,5.24)(56,5.51)(57,5.41)(58,5.6)(59,5.83)(60,4.35)
        (61,4.57)(62,3.92)(63,3.97)(64,4.13)(65,3.45)(66,3.63)(67,3.44)(68,3.5)(69,3.46)(70,3.4)
        (71,3.77)(72,4.62)(73,4.21)(74,4.2)(75,4.16)(76,4.08)(77,4.08)(78,4.18)(79,2.46)(80,2.6)
        };
            
    \addplot[
        color=black
        ]
        coordinates {
        (1,10)(2,10)(3,10)(4,10)(5,10)(6,10)(7,10)(8,10)(9,10)(10,10)
        (11,10)(12,10)(13,10)(14,10)(15,10)(16,10)(17,10)(18,10)(19,10)(20,10)
        (21,10)(22,10)(23,10)(24,10)(25,10)(26,10)(27,10)(28,10)(29,10)(30,10)
        (31,10)(32,10)(33,10)(34,10)(35,10)(36,10)(37,10)(38,10)(39,10)(40,10)
        (41,10)(42,10)(43,10)(44,10)(45,10)(46,10)(47,10)(48,10)(49,10)(50,10)
        (51,10)(52,10)(53,10)(54,10)(55,10)(56,10)(57,10)(58,10)(59,10)(60,10)
        (61,10)(62,10)(63,10)(64,10)(65,10)(66,10)(67,10)(68,10)(69,10)(70,10)
        (71,10)(72,10)(73,10)(74,10)(75,10)(76,10)(77,10)(78,10)(79,10)(80,10)
        };
                
        \legend{Non-Private FL, Local Learning, SMC, DPSGD, Hybrid-One, $\alpha$-CLDP-Fed, CLDP-Single Layer, CLDP-Basic, Baseline}
    \end{axis}
    \end{tikzpicture}}
    \caption{$\alpha$-CLDP-Fed compared to other FL methods.}
    \label{fig:by_layer}
    \vspace{-20pt}
\end{figure}
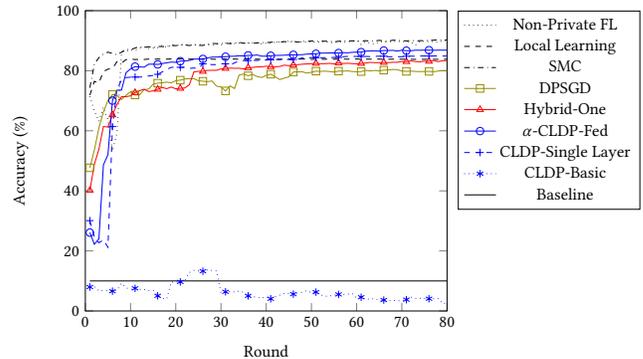
\subsection{Limited Updates with LDP-Fed}\label{subsec:update_size}
We first evaluate the effectiveness of LDP-Fed with $\alpha$-CLDP-Fed, a version of LDP-Fed with $\alpha$-CLDP in the LDP Module. The comparison study includes six existing federated learning scenarios and three FL settings using CLDP: CLDP-Basic, CLDP-Single Layer, and our recommended $\alpha$-CLDP-Fed. All private methods have a total privacy budget equivalent to $\alpha= 1.0$. 

Figure~\ref{fig:by_layer} reports the results. CLDP-Basic refers to a baseline implementation of $\alpha$-CLDP wherein participants provide updates at each round for all parameters in the DNN. Therefore, the budget $\alpha$ in CLDP-Basic must be divided amongst all the 29,034 parameters. As shown in Figure~\ref{fig:by_layer}, the CLDP-Basic displays the worst accuracy, below the random guest baseline of 10\%. This indicates that applying the privacy budget uniformly across all parameter updates can cause untenable loss of training accuracy for large, complex models. Instead, $\alpha$-CLDP-Fed presents a novel and intelligent algorithm for local differential privacy budget allocation and perturbation at each iteration throughout a FL workflow. In $\alpha$-CLDP-Fed, participants upload only a subset of the parameters at each round, resulting in a higher budget allocated to individual parameter uploads. We first describe CLDP-Single Layer. 

In CLDP-Single Layer, rather than sending a complete set $\theta$ of parameter updates at every round, each of the selected $k$ participants at round $i$ only perturbs and shares $\theta_i \subset \theta$ with the parameter server, where $\theta_i$ contains parameter updates for only a single layer of the DNN. The budget allocated to each parameter can then be increased to $\alpha_p = \frac{\alpha_i}{|\theta_i|}$ where $\alpha_i$ is the budget allocated to round $i$. Figure~\ref{fig:by_layer} shows that the CLDP-Single Layer algorithm significantly outperforms the CLDP-Basic algorithm and results in a final accuracy of 84.89\%. In CLDP-Single Layer, each layer is allocated an even number of rounds and each round an even slice of the budget. Specifically, given $\ell$ layers, the updates sent during the first $\frac{E}{\ell}$ rounds include only parameter updates for the parameters in the DNN output layer. During each subsequent $\frac{E}{\ell}$ set of rounds, updates are for parameters one layer backward in the network.

In contrast to CLDP-Single Layer, $\alpha$-CLDP-Fed allocates the number of rounds proportionate to the percentage of the model’s total parameters contained within that layer, i.e. for layer $i$, $E_i = \frac{|\theta_i|}{|\theta|}E$ total rounds are dedicated to updating parameters in layer $i$. A minimum of 1 round is reserved for each layer. The same backward stepping approach is used as in CLDP-Single Layer. In $\alpha$-CLDP-Fed the budget is also allocated proportionate to layer size. Figure~\ref{fig:by_layer} shows that $\alpha$-CLDP-Fed further improves the training accuracy of CLDP-Single Layer with the highest final accuracy among the privacy-preserving approaches with 86.85\% accuracy. Furthermore,  Figure~\ref{fig:by_layer} shows that both $\alpha$-CLDP-Fed and CLDP-Single Layer outperform the non-private Local Learning, DPSGD, and even Hybrid methods.

\begin{table}
    \centering
    \begin{tabular}{|c|c|c|}
    \hline
    \# of Cycles & Accuracy & Std Deviation \\
    \hline
    1           & 86.85\%           & 0.12          \\
    2           & 86.20\%           & 0.61          \\
    4           & 86.89\%           & 0.10          \\
    \textbf{5}  & \textbf{86.93\%}  & \textbf{0.12} \\
    10          & 86.30\%           & 0.24          \\
    16          & 85.28\%           & 0.11          \\
    \hline
    \end{tabular}
    \caption{Impact of introducing cycle-based approach in $\alpha$-CLDP-Fed. A minimum of $c'=$ number of cycles rounds allocated to each layer.}
    \label{tab:cycles}
    \vspace{-1.75\baselineskip}
\end{table}
\subsection{Impact of LDP-Fed Perturbation Cycles}\label{subsec:cycles}
In LDP-Fed we further introduces cycles to control when different parameter updates are shared with the parameter server. Each cycle is implemented in terms of iteration rounds. That is, let $c'=$ number of cycles. One cycle is then $\frac{E}{c'}$ rounds with each cycle being allocated $\frac{\alpha}{c'}$ of the privacy budget. Rounds and budget are then allocated within each cycle to individual layers according to the strategy in Section~\ref{subsec:update_size}. This allows layers to be revisited for updates within the training process. We report the impact of varying the number of cycles in $\alpha$-CLDP-Fed in Table 1. This set of experiments shows that setting the number of cycles to 5 will result in a high, stable accuracy of 86.93\% averaged across runs with a standard deviation of 0.12. In LDP-Fed, the default cycle value is set to 5. 

\begin{table*}[!h]
    \centering
    \begin{tabular}{|c|c|c|c|c|}
        \hline
        Privacy-Preserving          & Efficient & Locally Defined   & Protection from   & Handles Complex \\
        Federated Learning Method   &           & Privacy Guarantee & Inference Attacks & Models \\
        \hline
        SMC~\cite{bonawitz2017practical}
                        & \textcolor{red}{\large{\ding{55}}} & \textcolor{red}{\large{\ding{55}}} & \textcolor{BurntOrange}{\textbf{\Large{{\raise.17ex\hbox{\texttt{\textbf{$\sim$}}}}}}} &\textcolor{Green}{\large{\ding{51}}} \\
        \hline
        $\epsilon$-DP Parameter Sharing~\cite{shokri2015privacy}
                        & \textcolor{Green}{\large{\ding{51}}} & \textcolor{Green}{\large{\ding{51}}}
                        & \textcolor{BurntOrange}{\Large{{\raise.17ex\hbox{\texttt{\textbf{$\sim$}}}}}} & \textcolor{Green}{\large{\ding{51}}} \\
        \hline
        Local Optimizer~\cite{abadi2016deep}
                        & \textcolor{BurntOrange}{\textbf{\Large{{\raise.17ex\hbox{\texttt{\textbf{$\sim$}}}}}}} & \textcolor{Green}{\large{\ding{51}}} & \textcolor{Green}{\large{\ding{51}}} & \textcolor{red}{\large{\ding{55}}} \\
        \hline
        Hybrid-One~\cite{truex2019hybrid}
                        & \textcolor{red}{\large{\ding{55}}} & \textcolor{red}{\large{\ding{55}}} & \textcolor{Green}{\large{\ding{51}}} & \textcolor{BurntOrange}{\textbf{\Large{{\raise.17ex\hbox{\texttt{\textbf{$\sim$}}}}}}} \\
        \hline
        Continuous $\epsilon$-LDP~\cite{wang2019collecting}
                        & \textcolor{Green}{\large{\ding{51}}} & \textcolor{Green}{\large{\ding{51}}} & \textcolor{Green}{\large{\ding{51}}} & \textcolor{red}{\large{\ding{55}}} \\
        \hline
        LDP-Fed         &\textcolor{Green}{\large{\ding{51}}} &\textcolor{Green}{\large{\ding{51}}} & \textcolor{Green}{\large{\ding{51}}} & \textcolor{Green}{\large{\ding{51}}} \\
        \hline
    \end{tabular}
    \caption{Comparison of methods for private federated model training.}
    \label{tab:compare_checks}
    \vspace{-\baselineskip}
\end{table*}

\section{System Feature Comparison}\label{subsec:compare}
We have reported experimental comparison of our $\alpha$-CLDP-Fed method for privacy preserving federated learning with several representative approaches. We additionally provide a system feature comparison in Table~\ref{tab:compare_checks}; highlighting the value-added feature benefits of using LDP-Fed. First, LDP-Fed system does not require heavy cryptographic protocols which may not be suitable for edge devices engaged in FL. Second, LDP-Fed allows individual participants to locally define their own privacy level through the LDP Module. This is a valuable feature as previous work~\cite{truex2019effects,shokri2019privacy} has indicated that vulnerability to privacy attacks is not uniform and may be more acute for some participants' datasets, leading to a desire for a stricter privacy guarantee. Last, but not the least, LDP-Fed provides formal protection from known privacy inference attacks while demonstrating an ability to maintain good accuracy in the presence of large, complex models

\section{Related Work}\label{sec:related}
The LDP-Fed system relates to both FL and privacy-preserving ML. 

\textit{Federated Learning Approaches}. In~\cite{zhang2011distributed} authors propose a distribute data mining system with DP, but their results demonstrate a significant accuracy loss and the system requires a trusted aggregator to add the necessary noise. In~\cite{papernot:pate:2018}, while several ``teacher'' models are independently trained, a trusted aggregator must provide a DP query interface to a ``student'' model with unlabelled public data.~\cite{bonawitz2017practical} introduces cryptographic protocols to protect individual updates from being seen prior to aggregation, but leaves the aggregate updates and final predictive model vulnerable to inference attacks. Additional protocols allow users to leverage such cryptographic techniques to decrease the scale of noise~\cite{truex2019hybrid, dwork2006our, chase2017private}. These approaches require expensive cryptographic operations and either remove the ability of individual participants to identify privacy levels locally or demonstrate higher accuracy loss.

\textit{Privacy-Preserving ML}.~\cite{shokri2015privacy} similarly presents a distributed learning system using DP without a central trusted party. However, the DP guarantee is per-parameter and becomes meaningless for models with a large number of parameters.~\cite{wang2019collecting} also proposes an LDP protocol for multidimensional continuous data, however their experiments entailed 4 million users and $<$20 features for training smaller dimensional models.

\section{Conclusion}\label{sec:conclusion}
We have presented LDP-Fed, a novel federated learning approach with LDP. Our system allows participants to efficiently train complex models while providing formal privacy protection. The design of LDP-Fed has two unique features. First, it  enables participants to customize their LDP privacy budget locally according to their own preferences. Second, LDP-Fed implements a novel privacy preserving collaborative training approach towards utility-aware privacy perturbation to prevent uncontrolled noise from overwhelming the FL training algorithm in the presence of large, complex model parameter updates. The $\alpha$-CLDP-Fed algorithm design also exhibits a successful formal development of extending the traditional LDP theory, intended for single categorical values, to our LDP-Fed algorithm capable of handling high dimensional,  continuous, and large scale model parameter updates. We provide empirical and analytical comparison of LDP-Fed with the state-of-the-art privacy-preserving FL approaches in both accuracy and system features.

\begin{acks}
This research is partially sponsored by NSF CISE SaTC 1564097. The first author acknowledges an IBM PhD Fellowship Award and  the support from the Enterprise AI, Systems \& Solutions division led by Sandeep Gopisetty at IBM Almaden Research Center. CLDP-Fed is developed on top of the Ordinal-CLDP protocol whose implementation is a part of our CLDP release, publicly available at \url{https://github.com/git-disl/CLDP}. Any opinions, findings, and conclusions or recommendations expressed in this material are those of the author(s) and do not necessarily reflect the views of the National Science Foundation or other funding agencies and companies mentioned above. 
\end{acks}

\bibliographystyle{ACM-Reference-Format}
\bibliography{edgesys}


\begin{thebibliography}{28}


\ifx \showCODEN    \undefined \def \showCODEN     #1{\unskip}     \fi
\ifx \showDOI      \undefined \def \showDOI       #1{#1}\fi
\ifx \showISBNx    \undefined \def \showISBNx     #1{\unskip}     \fi
\ifx \showISBNxiii \undefined \def \showISBNxiii  #1{\unskip}     \fi
\ifx \showISSN     \undefined \def \showISSN      #1{\unskip}     \fi
\ifx \showLCCN     \undefined \def \showLCCN      #1{\unskip}     \fi
\ifx \shownote     \undefined \def \shownote      #1{#1}          \fi
\ifx \showarticletitle \undefined \def \showarticletitle #1{#1}   \fi
\ifx \showURL      \undefined \def \showURL       {\relax}        \fi
\providecommand\bibfield[2]{#2}
\providecommand\bibinfo[2]{#2}
\providecommand\natexlab[1]{#1}
\providecommand\showeprint[2][]{arXiv:#2}

\bibitem[\protect\citeauthoryear{Abadi, Barham, Chen, Chen, Davis, Dean, Devin,
  Ghemawat, Irving, Isard, et~al\mbox{.}}{Abadi et~al\mbox{.}}{2016a}]%
        {abadi2016tensorflow}
\bibfield{author}{\bibinfo{person}{Mart{\'\i}n Abadi}, \bibinfo{person}{Paul
  Barham}, \bibinfo{person}{Jianmin Chen}, \bibinfo{person}{Zhifeng Chen},
  \bibinfo{person}{Andy Davis}, \bibinfo{person}{Jeffrey Dean},
  \bibinfo{person}{Matthieu Devin}, \bibinfo{person}{Sanjay Ghemawat},
  \bibinfo{person}{Geoffrey Irving}, \bibinfo{person}{Michael Isard},
  {et~al\mbox{.}}} \bibinfo{year}{2016}\natexlab{a}.
\newblock \showarticletitle{Tensorflow: A system for large-scale machine
  learning}. In \bibinfo{booktitle}{\emph{12th $\{$USENIX$\}$ Symposium on
  Operating Systems Design and Implementation ($\{$OSDI$\}$ 16)}}.
  \bibinfo{pages}{265--283}.
\newblock


\bibitem[\protect\citeauthoryear{Abadi, Chu, Goodfellow, McMahan, Mironov,
  Talwar, and Zhang}{Abadi et~al\mbox{.}}{2016b}]%
        {abadi2016deep}
\bibfield{author}{\bibinfo{person}{Martin Abadi}, \bibinfo{person}{Andy Chu},
  \bibinfo{person}{Ian Goodfellow}, \bibinfo{person}{H~Brendan McMahan},
  \bibinfo{person}{Ilya Mironov}, \bibinfo{person}{Kunal Talwar}, {and}
  \bibinfo{person}{Li Zhang}.} \bibinfo{year}{2016}\natexlab{b}.
\newblock \showarticletitle{Deep learning with differential privacy}. In
  \bibinfo{booktitle}{\emph{Proceedings of the 2016 ACM SIGSAC Conference on
  Computer and Communications Security}}. \bibinfo{pages}{308--318}.
\newblock


\bibitem[\protect\citeauthoryear{Act}{Act}{1996}]%
        {act1996health}
\bibfield{author}{\bibinfo{person}{Accountability Act}.}
  \bibinfo{year}{1996}\natexlab{}.
\newblock \showarticletitle{Health insurance portability and accountability act
  of 1996}.
\newblock \bibinfo{journal}{\emph{Public law}}  \bibinfo{volume}{104}
  (\bibinfo{year}{1996}), \bibinfo{pages}{191}.
\newblock


\bibitem[\protect\citeauthoryear{Beimel, Brenner, Kasiviswanathan, and
  Nissim}{Beimel et~al\mbox{.}}{2014}]%
        {beimel2014bounds}
\bibfield{author}{\bibinfo{person}{Amos Beimel}, \bibinfo{person}{Hai Brenner},
  \bibinfo{person}{Shiva~Prasad Kasiviswanathan}, {and} \bibinfo{person}{Kobbi
  Nissim}.} \bibinfo{year}{2014}\natexlab{}.
\newblock \showarticletitle{Bounds on the sample complexity for private
  learning and private data release}.
\newblock \bibinfo{journal}{\emph{Machine learning}} \bibinfo{volume}{94},
  \bibinfo{number}{3} (\bibinfo{year}{2014}), \bibinfo{pages}{401--437}.
\newblock


\bibitem[\protect\citeauthoryear{Bonawitz, Eichner, Grieskamp, Huba, Ingerman,
  Ivanov, Kiddon, Konecny, Mazzocchi, McMahan, et~al\mbox{.}}{Bonawitz
  et~al\mbox{.}}{2019}]%
        {bonawitz2019towards}
\bibfield{author}{\bibinfo{person}{Keith Bonawitz}, \bibinfo{person}{Hubert
  Eichner}, \bibinfo{person}{Wolfgang Grieskamp}, \bibinfo{person}{Dzmitry
  Huba}, \bibinfo{person}{Alex Ingerman}, \bibinfo{person}{Vladimir Ivanov},
  \bibinfo{person}{Chloe Kiddon}, \bibinfo{person}{Jakub Konecny},
  \bibinfo{person}{Stefano Mazzocchi}, \bibinfo{person}{H~Brendan McMahan},
  {et~al\mbox{.}}} \bibinfo{year}{2019}\natexlab{}.
\newblock \showarticletitle{Towards federated learning at scale: System
  design}.
\newblock \bibinfo{journal}{\emph{arXiv preprint arXiv:1902.01046}}
  (\bibinfo{year}{2019}).
\newblock


\bibitem[\protect\citeauthoryear{Bonawitz, Ivanov, Kreuter, Marcedone, McMahan,
  Patel, Ramage, Segal, and Seth}{Bonawitz et~al\mbox{.}}{2017}]%
        {bonawitz2017practical}
\bibfield{author}{\bibinfo{person}{Keith Bonawitz}, \bibinfo{person}{Vladimir
  Ivanov}, \bibinfo{person}{Ben Kreuter}, \bibinfo{person}{Antonio Marcedone},
  \bibinfo{person}{H~Brendan McMahan}, \bibinfo{person}{Sarvar Patel},
  \bibinfo{person}{Daniel Ramage}, \bibinfo{person}{Aaron Segal}, {and}
  \bibinfo{person}{Karn Seth}.} \bibinfo{year}{2017}\natexlab{}.
\newblock \showarticletitle{Practical secure aggregation for privacy-preserving
  machine learning}. In \bibinfo{booktitle}{\emph{Proceedings of the 2017 ACM
  SIGSAC Conference on Computer and Communications Security}}. ACM,
  \bibinfo{pages}{1175--1191}.
\newblock


\bibitem[\protect\citeauthoryear{Chase, Gilad-Bachrach, Laine, Lauter, and
  Rindal}{Chase et~al\mbox{.}}{2017}]%
        {chase2017private}
\bibfield{author}{\bibinfo{person}{Melissa Chase}, \bibinfo{person}{Ran
  Gilad-Bachrach}, \bibinfo{person}{Kim Laine}, \bibinfo{person}{Kristin~E
  Lauter}, {and} \bibinfo{person}{Peter Rindal}.}
  \bibinfo{year}{2017}\natexlab{}.
\newblock \showarticletitle{Private Collaborative Neural Network Learning.}
\newblock \bibinfo{journal}{\emph{IACR Cryptology ePrint Archive}}
  \bibinfo{volume}{2017} (\bibinfo{year}{2017}), \bibinfo{pages}{762}.
\newblock


\bibitem[\protect\citeauthoryear{Collobert, Kavukcuoglu, and Farabet}{Collobert
  et~al\mbox{.}}{2011}]%
        {collobert2011torch7}
\bibfield{author}{\bibinfo{person}{Ronan Collobert}, \bibinfo{person}{Koray
  Kavukcuoglu}, {and} \bibinfo{person}{Cl{\'e}ment Farabet}.}
  \bibinfo{year}{2011}\natexlab{}.
\newblock \showarticletitle{Torch7: A matlab-like environment for machine
  learning}. In \bibinfo{booktitle}{\emph{BigLearn, NIPS workshop}}.
\newblock


\bibitem[\protect\citeauthoryear{Ding, Kulkarni, and Yekhanin}{Ding
  et~al\mbox{.}}{2017}]%
        {ding2017collecting}
\bibfield{author}{\bibinfo{person}{Bolin Ding}, \bibinfo{person}{Janardhan
  Kulkarni}, {and} \bibinfo{person}{Sergey Yekhanin}.}
  \bibinfo{year}{2017}\natexlab{}.
\newblock \showarticletitle{Collecting telemetry data privately}. In
  \bibinfo{booktitle}{\emph{Advances in Neural Information Processing
  Systems}}. \bibinfo{pages}{3571--3580}.
\newblock


\bibitem[\protect\citeauthoryear{Dwork, Kenthapadi, McSherry, Mironov, and
  Naor}{Dwork et~al\mbox{.}}{2006}]%
        {dwork2006our}
\bibfield{author}{\bibinfo{person}{Cynthia Dwork}, \bibinfo{person}{Krishnaram
  Kenthapadi}, \bibinfo{person}{Frank McSherry}, \bibinfo{person}{Ilya
  Mironov}, {and} \bibinfo{person}{Moni Naor}.}
  \bibinfo{year}{2006}\natexlab{}.
\newblock \showarticletitle{Our data, ourselves: Privacy via distributed noise
  generation}. In \bibinfo{booktitle}{\emph{Annual International Conference on
  the Theory and Applications of Cryptographic Techniques}}. Springer,
  \bibinfo{pages}{486--503}.
\newblock


\bibitem[\protect\citeauthoryear{Erlingsson, Pihur, and Korolova}{Erlingsson
  et~al\mbox{.}}{2014}]%
        {erlingsson2014rappor}
\bibfield{author}{\bibinfo{person}{{\'U}lfar Erlingsson},
  \bibinfo{person}{Vasyl Pihur}, {and} \bibinfo{person}{Aleksandra Korolova}.}
  \bibinfo{year}{2014}\natexlab{}.
\newblock \showarticletitle{Rappor: Randomized aggregatable privacy-preserving
  ordinal response}. In \bibinfo{booktitle}{\emph{Proceedings of the 2014 ACM
  SIGSAC conference on computer and communications security}}.
  \bibinfo{pages}{1054--1067}.
\newblock


\bibitem[\protect\citeauthoryear{Gursoy, Tamersoy, Truex, Wei, and Liu}{Gursoy
  et~al\mbox{.}}{2019}]%
        {gursoy2019secure}
\bibfield{author}{\bibinfo{person}{Mehmet~Emre Gursoy}, \bibinfo{person}{Acar
  Tamersoy}, \bibinfo{person}{Stacey Truex}, \bibinfo{person}{Wenqi Wei}, {and}
  \bibinfo{person}{Ling Liu}.} \bibinfo{year}{2019}\natexlab{}.
\newblock \showarticletitle{Secure and utility-aware data collection with
  condensed local differential privacy}.
\newblock \bibinfo{journal}{\emph{IEEE Transactions on Dependable and Secure
  Computing}} (\bibinfo{year}{2019}).
\newblock


\bibitem[\protect\citeauthoryear{Ierusalimschy, De~Figueiredo, and
  Filho}{Ierusalimschy et~al\mbox{.}}{1996}]%
        {ierusalimschy1996lua}
\bibfield{author}{\bibinfo{person}{Roberto Ierusalimschy},
  \bibinfo{person}{Luiz~Henrique De~Figueiredo}, {and}
  \bibinfo{person}{Waldemar~Celes Filho}.} \bibinfo{year}{1996}\natexlab{}.
\newblock \showarticletitle{Lua—an extensible extension language}.
\newblock \bibinfo{journal}{\emph{Software: Practice and Experience}}
  \bibinfo{volume}{26}, \bibinfo{number}{6} (\bibinfo{year}{1996}),
  \bibinfo{pages}{635--652}.
\newblock


\bibitem[\protect\citeauthoryear{Kasiviswanathan, Lee, Nissim, Raskhodnikova,
  and Smith}{Kasiviswanathan et~al\mbox{.}}{2011}]%
        {kasiviswanathan2011can}
\bibfield{author}{\bibinfo{person}{Shiva~Prasad Kasiviswanathan},
  \bibinfo{person}{Homin~K Lee}, \bibinfo{person}{Kobbi Nissim},
  \bibinfo{person}{Sofya Raskhodnikova}, {and} \bibinfo{person}{Adam Smith}.}
  \bibinfo{year}{2011}\natexlab{}.
\newblock \showarticletitle{What can we learn privately?}
\newblock \bibinfo{journal}{\emph{SIAM J. Comput.}} \bibinfo{volume}{40},
  \bibinfo{number}{3} (\bibinfo{year}{2011}), \bibinfo{pages}{793--826}.
\newblock


\bibitem[\protect\citeauthoryear{Mathews and Bowman}{Mathews and
  Bowman}{2018}]%
        {mathews2018california}
\bibfield{author}{\bibinfo{person}{KJ Mathews} {and} \bibinfo{person}{CM
  Bowman}.} \bibinfo{year}{2018}\natexlab{}.
\newblock \bibinfo{title}{The California Consumer Privacy Act of 2018}.
\newblock
\newblock


\bibitem[\protect\citeauthoryear{Nasr, Shokri, and Houmansadr}{Nasr
  et~al\mbox{.}}{2019}]%
        {nasr2018comprehensive}
\bibfield{author}{\bibinfo{person}{Milad Nasr}, \bibinfo{person}{Reza Shokri},
  {and} \bibinfo{person}{Amir Houmansadr}.} \bibinfo{year}{2019}\natexlab{}.
\newblock \showarticletitle{Comprehensive Privacy Analysis of Deep Learning:
  Stand-alone and Federated Learning under Passive and Active White-box
  Inference Attacks}. In \bibinfo{booktitle}{\emph{Security and Privacy (SP),
  2019 IEEE Symposium on}}.
\newblock


\bibitem[\protect\citeauthoryear{Papernot, Song, Mironov, Raghunathan, Talwar,
  and Erlingsson}{Papernot et~al\mbox{.}}{2018}]%
        {papernot:pate:2018}
\bibfield{author}{\bibinfo{person}{Nicolas Papernot}, \bibinfo{person}{Shuang
  Song}, \bibinfo{person}{Ilya Mironov}, \bibinfo{person}{Ananth Raghunathan},
  \bibinfo{person}{Kunal Talwar}, {and} \bibinfo{person}{{\'U}lfar
  Erlingsson}.} \bibinfo{year}{2018}\natexlab{}.
\newblock \showarticletitle{Scalable Private Learning with PATE}.
\newblock \bibinfo{journal}{\emph{arXiv preprint arXiv:1802.08908}}
  (\bibinfo{year}{2018}).
\newblock


\bibitem[\protect\citeauthoryear{Regulation}{Regulation}{2016}]%
        {regulation2016regulation}
\bibfield{author}{\bibinfo{person}{General Data~Protection Regulation}.}
  \bibinfo{year}{2016}\natexlab{}.
\newblock \showarticletitle{Regulation (EU) 2016/679 of the European Parliament
  and of the Council of 27 April 2016 on the protection of natural persons with
  regard to the processing of personal data and on the free movement of such
  data, and repealing Directive 95/46}.
\newblock \bibinfo{journal}{\emph{Official Journal of the European Union (OJ)}}
  \bibinfo{volume}{59}, \bibinfo{number}{1-88} (\bibinfo{year}{2016}),
  \bibinfo{pages}{294}.
\newblock


\bibitem[\protect\citeauthoryear{Shokri and Shmatikov}{Shokri and
  Shmatikov}{2015}]%
        {shokri2015privacy}
\bibfield{author}{\bibinfo{person}{Reza Shokri} {and} \bibinfo{person}{Vitaly
  Shmatikov}.} \bibinfo{year}{2015}\natexlab{}.
\newblock \showarticletitle{Privacy-preserving deep learning}. In
  \bibinfo{booktitle}{\emph{Proceedings of the 22nd ACM SIGSAC conference on
  computer and communications security}}. \bibinfo{pages}{1310--1321}.
\newblock


\bibitem[\protect\citeauthoryear{Shokri, Strobel, and Zick}{Shokri
  et~al\mbox{.}}{2019}]%
        {shokri2019privacy}
\bibfield{author}{\bibinfo{person}{Reza Shokri}, \bibinfo{person}{Martin
  Strobel}, {and} \bibinfo{person}{Yair Zick}.}
  \bibinfo{year}{2019}\natexlab{}.
\newblock \showarticletitle{Privacy risks of explaining machine learning
  models}.
\newblock \bibinfo{journal}{\emph{arXiv preprint arXiv:1907.00164}}
  (\bibinfo{year}{2019}).
\newblock


\bibitem[\protect\citeauthoryear{Shokri, Stronati, Song, and Shmatikov}{Shokri
  et~al\mbox{.}}{2017}]%
        {shokri2017membership}
\bibfield{author}{\bibinfo{person}{Reza Shokri}, \bibinfo{person}{Marco
  Stronati}, \bibinfo{person}{Congzheng Song}, {and} \bibinfo{person}{Vitaly
  Shmatikov}.} \bibinfo{year}{2017}\natexlab{}.
\newblock \showarticletitle{Membership inference attacks against machine
  learning models}. In \bibinfo{booktitle}{\emph{2017 IEEE Symposium on
  Security and Privacy (SP)}}. IEEE, \bibinfo{pages}{3--18}.
\newblock


\bibitem[\protect\citeauthoryear{Thakurta, Vyrros, Vaishampayan, Kapoor,
  Freudinger, Prakash, Legendre, and Duplinsky}{Thakurta et~al\mbox{.}}{2017}]%
        {thakurta2017emoji}
\bibfield{author}{\bibinfo{person}{Abhradeep~Guha Thakurta},
  \bibinfo{person}{Andrew~H Vyrros}, \bibinfo{person}{Umesh~S Vaishampayan},
  \bibinfo{person}{Gaurav Kapoor}, \bibinfo{person}{Julien Freudinger},
  \bibinfo{person}{Vipul~Ved Prakash}, \bibinfo{person}{Arnaud Legendre}, {and}
  \bibinfo{person}{Steven Duplinsky}.} \bibinfo{year}{2017}\natexlab{}.
\newblock \bibinfo{title}{Emoji frequency detection and deep link frequency}.
\newblock
\newblock
\newblock
\shownote{US Patent 9,705,908.}


\bibitem[\protect\citeauthoryear{Truex, Baracaldo, Anwar, Steinke, Ludwig,
  Zhang, and Zhou}{Truex et~al\mbox{.}}{2019a}]%
        {truex2019hybrid}
\bibfield{author}{\bibinfo{person}{Stacey Truex}, \bibinfo{person}{Nathalie
  Baracaldo}, \bibinfo{person}{Ali Anwar}, \bibinfo{person}{Thomas Steinke},
  \bibinfo{person}{Heiko Ludwig}, \bibinfo{person}{Rui Zhang}, {and}
  \bibinfo{person}{Yi Zhou}.} \bibinfo{year}{2019}\natexlab{a}.
\newblock \showarticletitle{A hybrid approach to privacy-preserving federated
  learning}. In \bibinfo{booktitle}{\emph{Proceedings of the 12th ACM Workshop
  on Artificial Intelligence and Security}}. \bibinfo{pages}{1--11}.
\newblock


\bibitem[\protect\citeauthoryear{Truex, Liu, Gursoy, Wei, and Yu}{Truex
  et~al\mbox{.}}{2019b}]%
        {truex2019effects}
\bibfield{author}{\bibinfo{person}{Stacey Truex}, \bibinfo{person}{Ling Liu},
  \bibinfo{person}{Mehmet~Emre Gursoy}, \bibinfo{person}{Wenqi Wei}, {and}
  \bibinfo{person}{Lei Yu}.} \bibinfo{year}{2019}\natexlab{b}.
\newblock \showarticletitle{Effects of Differential Privacy and Data Skewness
  on Membership Inference Vulnerability}.
\newblock \bibinfo{journal}{\emph{arXiv preprint arXiv:1911.09777}}
  (\bibinfo{year}{2019}).
\newblock


\bibitem[\protect\citeauthoryear{Truex, Liu, Gursoy, Yu, and Wei}{Truex
  et~al\mbox{.}}{2019c}]%
        {truex2019demystifying}
\bibfield{author}{\bibinfo{person}{Stacey Truex}, \bibinfo{person}{Ling Liu},
  \bibinfo{person}{Mehmet~Emre Gursoy}, \bibinfo{person}{Lei Yu}, {and}
  \bibinfo{person}{Wenqi Wei}.} \bibinfo{year}{2019}\natexlab{c}.
\newblock \showarticletitle{Demystifying membership inference attacks in
  machine learning as a service}.
\newblock \bibinfo{journal}{\emph{IEEE Transactions on Services Computing}}
  (\bibinfo{year}{2019}).
\newblock


\bibitem[\protect\citeauthoryear{Wang, Xiao, Yang, Zhao, Hui, Shin, Shin, and
  Yu}{Wang et~al\mbox{.}}{2019}]%
        {wang2019collecting}
\bibfield{author}{\bibinfo{person}{Ning Wang}, \bibinfo{person}{Xiaokui Xiao},
  \bibinfo{person}{Yin Yang}, \bibinfo{person}{Jun Zhao},
  \bibinfo{person}{Siu~Cheung Hui}, \bibinfo{person}{Hyejin Shin},
  \bibinfo{person}{Junbum Shin}, {and} \bibinfo{person}{Ge Yu}.}
  \bibinfo{year}{2019}\natexlab{}.
\newblock \showarticletitle{Collecting and analyzing multidimensional data with
  local differential privacy}. In \bibinfo{booktitle}{\emph{2019 IEEE 35th
  International Conference on Data Engineering (ICDE)}}. IEEE,
  \bibinfo{pages}{638--649}.
\newblock


\bibitem[\protect\citeauthoryear{Xiao, Rasul, and Vollgraf}{Xiao
  et~al\mbox{.}}{2017}]%
        {xiao2017/online}
\bibfield{author}{\bibinfo{person}{Han Xiao}, \bibinfo{person}{Kashif Rasul},
  {and} \bibinfo{person}{Roland Vollgraf}.} \bibinfo{year}{2017}\natexlab{}.
\newblock \bibinfo{booktitle}{\emph{Fashion-MNIST: a Novel Image Dataset for
  Benchmarking Machine Learning Algorithms}}.
\newblock
\showeprint[arXiv]{cs.LG/cs.LG/1708.07747}


\bibitem[\protect\citeauthoryear{Zhang, Li, and Lou}{Zhang
  et~al\mbox{.}}{2011}]%
        {zhang2011distributed}
\bibfield{author}{\bibinfo{person}{Ning Zhang}, \bibinfo{person}{Ming Li},
  {and} \bibinfo{person}{Wenjing Lou}.} \bibinfo{year}{2011}\natexlab{}.
\newblock \showarticletitle{Distributed data mining with differential privacy}.
  In \bibinfo{booktitle}{\emph{Communications (ICC), 2011 IEEE International
  Conference on}}. IEEE, \bibinfo{pages}{1--5}.
\newblock


\end{thebibliography}

\end{document}